\documentclass{article}

\usepackage[preprint]{neurips_2025}

\usepackage[utf8]{inputenc} 
\usepackage[T1]{fontenc}    
\usepackage{hyperref}       
\usepackage{url}            
\usepackage{booktabs}       
\usepackage{amsfonts}       
\usepackage{amsmath}
\usepackage{nicefrac}       
\usepackage{microtype}      
\usepackage{xcolor}         

\usepackage{graphicx}
\usepackage{subcaption}

\title{AttnMod: Attention-Based New Art Styles}

\author{%
  Shih-Chieh Su \\
  San Diego, CA 92130 \\
  \texttt{jessysu@gmail.com} \\
}

\begin{document}

\maketitle

\begin{figure}[h]
    \centering
    \begin{subfigure}[b]{\textwidth}
        \includegraphics[width=1.0\textwidth]{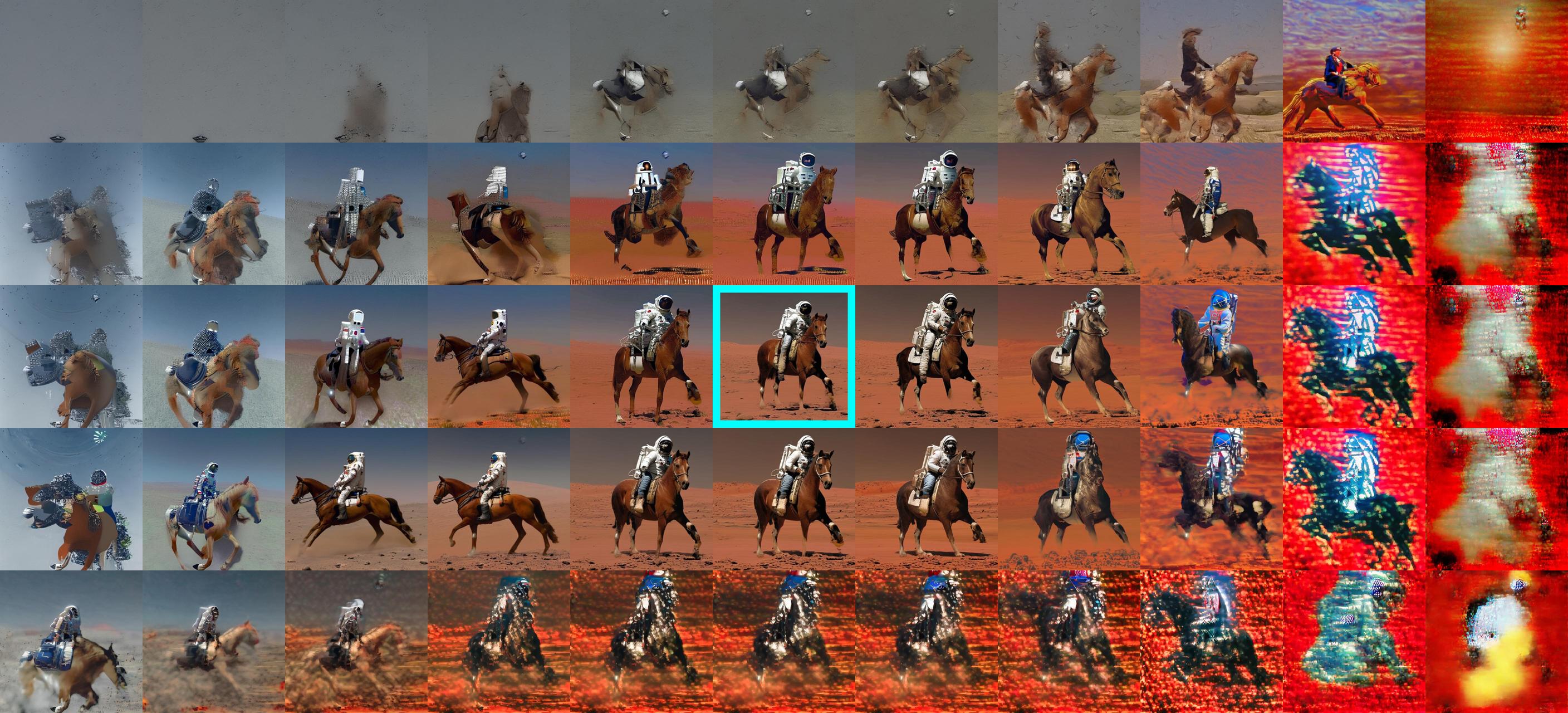}
    \end{subfigure}
    \caption{Generated outputs using AttnMod applied to a single attention block of a diffusion model. All images share the same prompt, seed, model, and denoising settings. Variations in style arise solely from differences in how attention is modulated during denoising. The center image shows the output without AttnMod for comparison.}

    \label{fig:teaser}
\end{figure}

\begin{abstract}
Despite using identical generation parameters—prompt, seed, model, scheduler, and denoising steps—diffusion models can produce visually distinct outputs when attention is modified during inference. Figure~\ref{fig:teaser} demonstrates this effect. We introduce AttnMod, a lightweight intervention that perturbs cross-attention within the denoising process to synthesize new visual styles. AttnMod requires no model retraining or prompt engineering and operates directly on pre-trained diffusion models. It enables the discovery of coherent, unpromptable styles, expanding the creative capacity of text-to-image systems.
\end{abstract}

\section{Introduction}

\begin{figure}[ht]
    \centering
    \begin{subfigure}[b]{0.65\textwidth}
        \includegraphics[width=\textwidth]{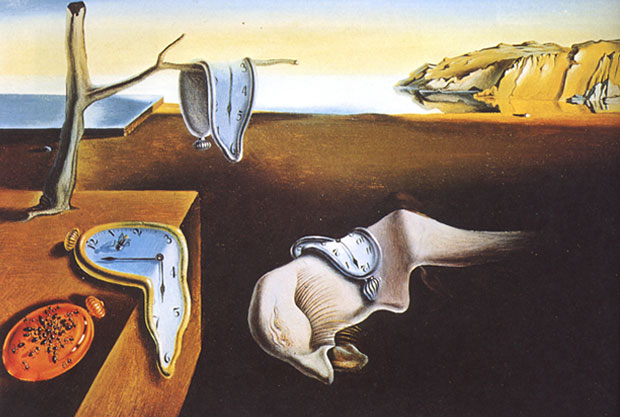}
        \caption{\textit{Salvador Dalí, The Persistence of Memory} (1931)}
    \end{subfigure}
    \begin{subfigure}[b]{0.34\textwidth}
        \includegraphics[width=\textwidth]{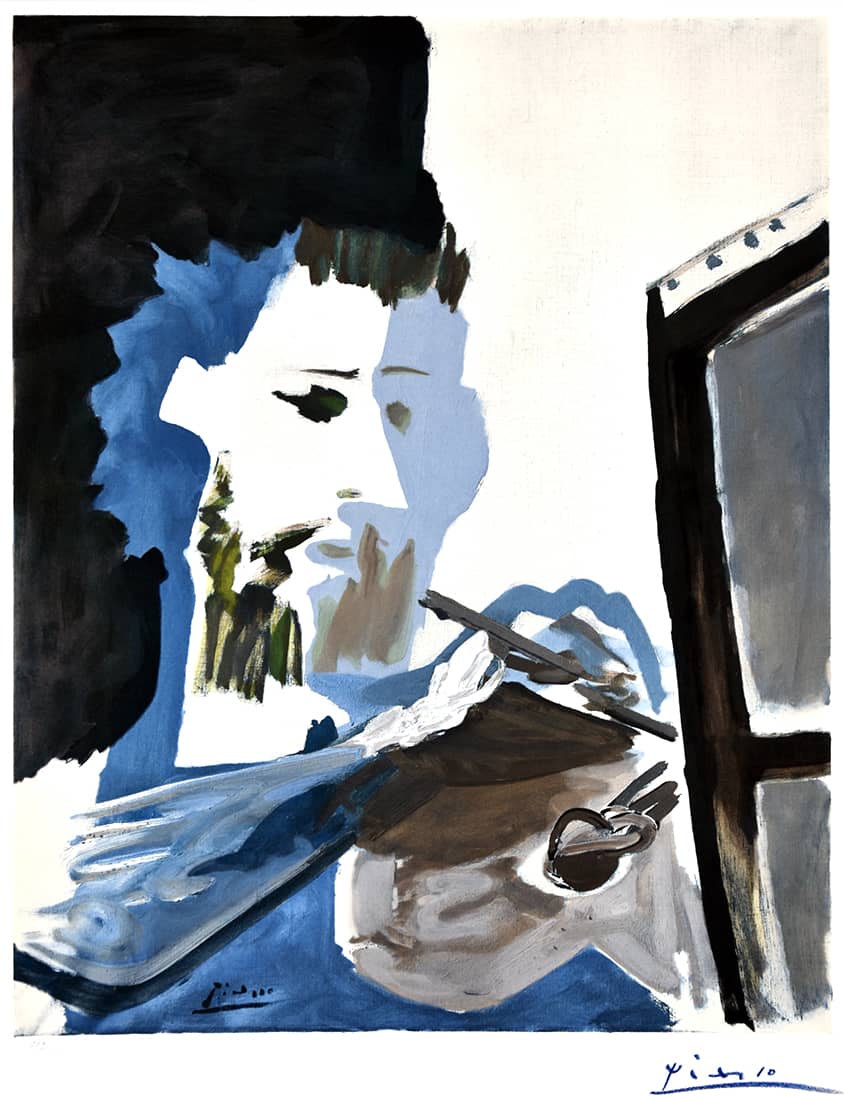}
        \caption{\textit{Pablo Picasso, Le Peintre} (1963)}
    \end{subfigure}
    \caption{Avant garde art concepts are often difficult to describe or express through text prompts alone.}
    \label{fig:unpromptable}
\end{figure}

Style transfer aims to render the content of one image in the artistic style of another, producing an image that preserves the semantic structure of the content while adopting the visual appearance of the style~\cite{gatys2016image}. With the rise of diffusion based generative models, which offer high visual fidelity~\cite{ho2020denoising}, improved sampling efficiency~\cite{song2020denoising,karras2022elucidating}, and compact latent space representations~\cite{rombach2022high}, recent research has explored their potential for style transfer~\cite{sohn2023styledrop,hertz2024style}. Despite these advances, the creation of entirely new artistic styles that are not easily described or evoked by text prompts or reference images remains largely unexplored. Figure~\ref{fig:unpromptable} illustrates examples of such hard to prompt art styles in existing avant garde works.

\begin{figure}[ht]
    \centering
    \begin{subfigure}[b]{\textwidth}
        \includegraphics[width=1.0\textwidth]{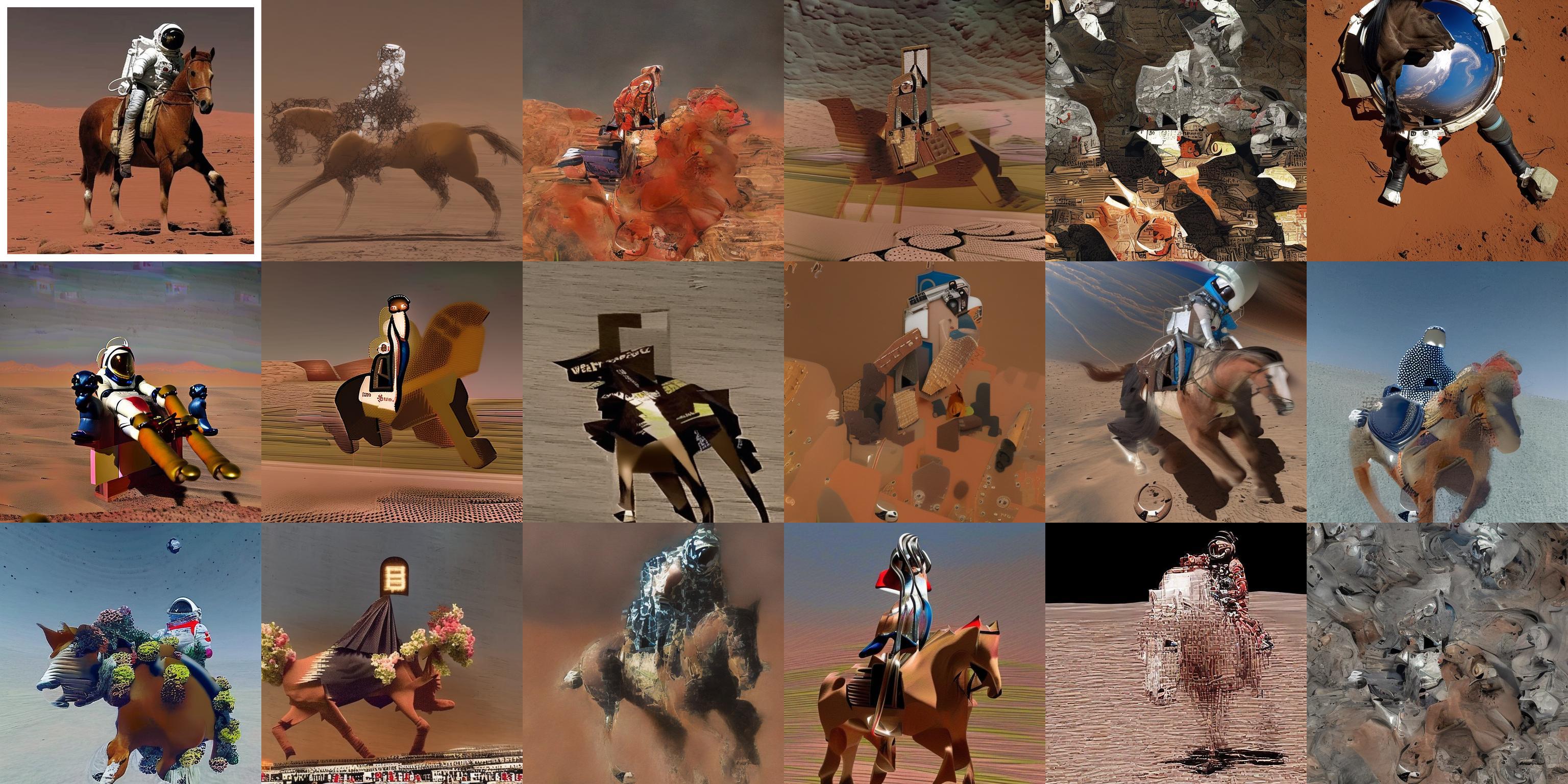}
    \end{subfigure}
    \caption{Images generated with AttnMod using the same configuration as in Figure~\ref{fig:teaser}. All inputs including prompt, seed, model, scheduler, and denoising steps are held constant. The only variation is how attention is modified. The top left image shows the baseline result without AttnMod.}
    \label{fig:picked_output}
\end{figure}

In this work, we explore the role of attention in artistic style formation by modifying the attention layers within the U-Net~\cite{ronneberger2015u} of the text-to-image (T2I) Stable Diffusion model~\cite{rombach2022high}. Using a fixed generation configuration, which includes prompt, model, scheduler, seed, and number of denoising steps, we then introduce controlled changes to the attention mechanism and observe substantial variations in the resulting visual style. An example is shown in Figure~\ref{fig:picked_output}, where the only difference between images is the way attention is adjusted during denoising.

We analyze how style characteristics evolve by either varying the seed while keeping the prompt fixed, or varying the prompt while keeping the seed fixed. Our study extends across different attention blocks and checkpoints of diffusion models. The proposed method, AttnMod, introduces novel styles directly from pretrained models without training or fine tuning. It offers a flexible mechanism for artistic control, especially in situations where the desired visual concept cannot be clearly expressed using prompts or example images.

\section{Methods}

Consider a human artist interpreting a scene or a concept, then reimagining it as a new painting. The artist may choose to emphasize particular visual features, distort object boundaries, adjust color distribution, or render new elements into the scene. This intuitive and creative reinterpretation process inspires our approach: to simulate such stylistic intent through controlled modification---i.e., modulation---of the attention mechanism in diffusion models.

\begin{figure}[ht]
    \centering
    \begin{subfigure}[b]{\textwidth}
        \includegraphics[width=1.0\textwidth]{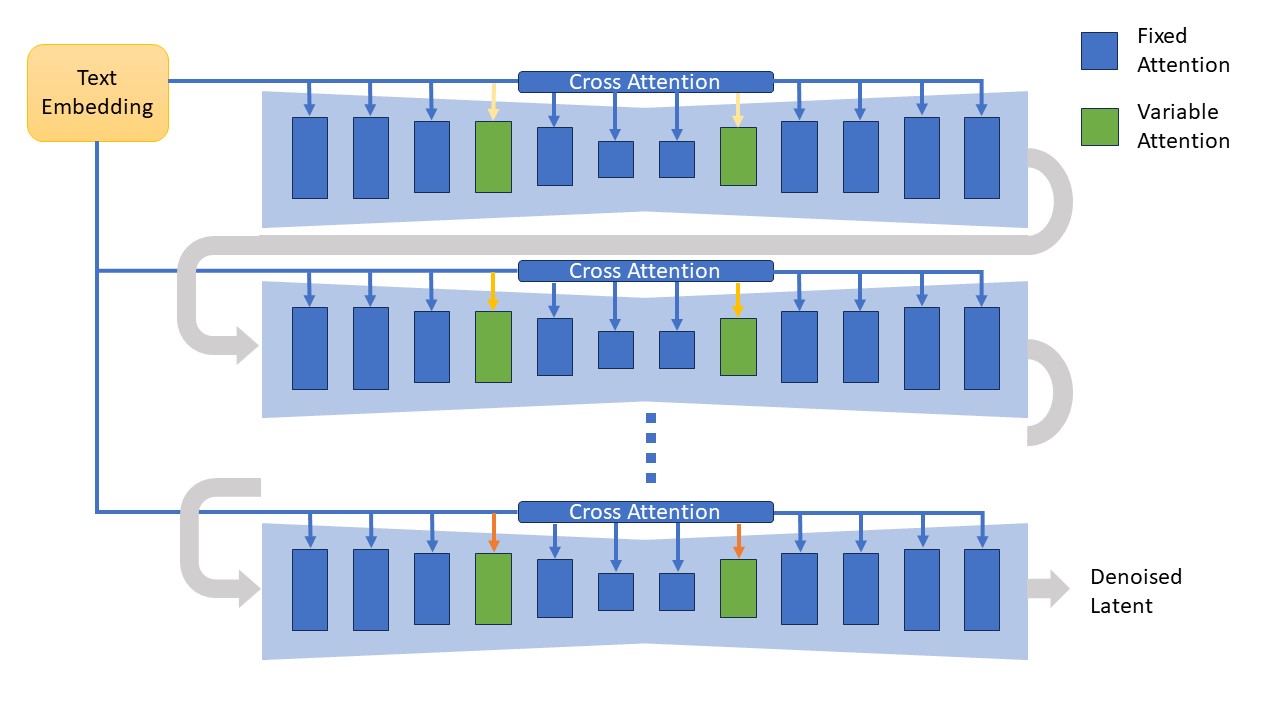}
    \end{subfigure}
    \caption{Overview of the AttnMod process.}
    \label{fig:attnmod_flow}
\end{figure}

We propose \textbf{AttnMod}, a method that alters the behavior of cross-attention in the U-Net of a pretrained T2I diffusion model. As shown in Figure~\ref{fig:attnmod_flow}, AttnMod selectively modifies attention strength over the course of the denoising process, allowing the model to generate novel visual styles without changing the prompt, seed, or model weights.

\subsection{Attention in Diffusion Models}

Diffusion-based image generation starts from pure noise and iteratively denoises the latent representation using a U-Net~\cite{ronneberger2015u}, guided by a text prompt. The prompt is embedded and injected into the denoising loop via cross-attention layers in the U-Net. During training, these attention layers learn how to condition the denoising process based on the semantic meaning of the prompt, ultimately guiding the model toward the target image~\cite{rombach2022high}.

To manipulate the strength of conditioning at inference time, we introduce two scalar parameters: the \textit{attention multiplier} $\alpha$ and the \textit{in-loop change rate} $\delta$. These parameters control the scaling of cross-attention values within a specific attention block of the U-Net. The attention at each step $t$ is scaled by:
\[
\text{Attention}(t) = \alpha + t \cdot \delta
\]

\subsection{AttnMod Scans}

To study how different attention settings affect the generated image, we conduct a two-dimensional scan over $\alpha$ and $\delta$ for a selected cross-attention block, while leaving all others unmodified (i.e., attention multiplier set to 1.0).

\begin{figure}[ht]
    \centering
    \begin{subfigure}[b]{0.92\textwidth}
        \includegraphics[width=1.0\textwidth]{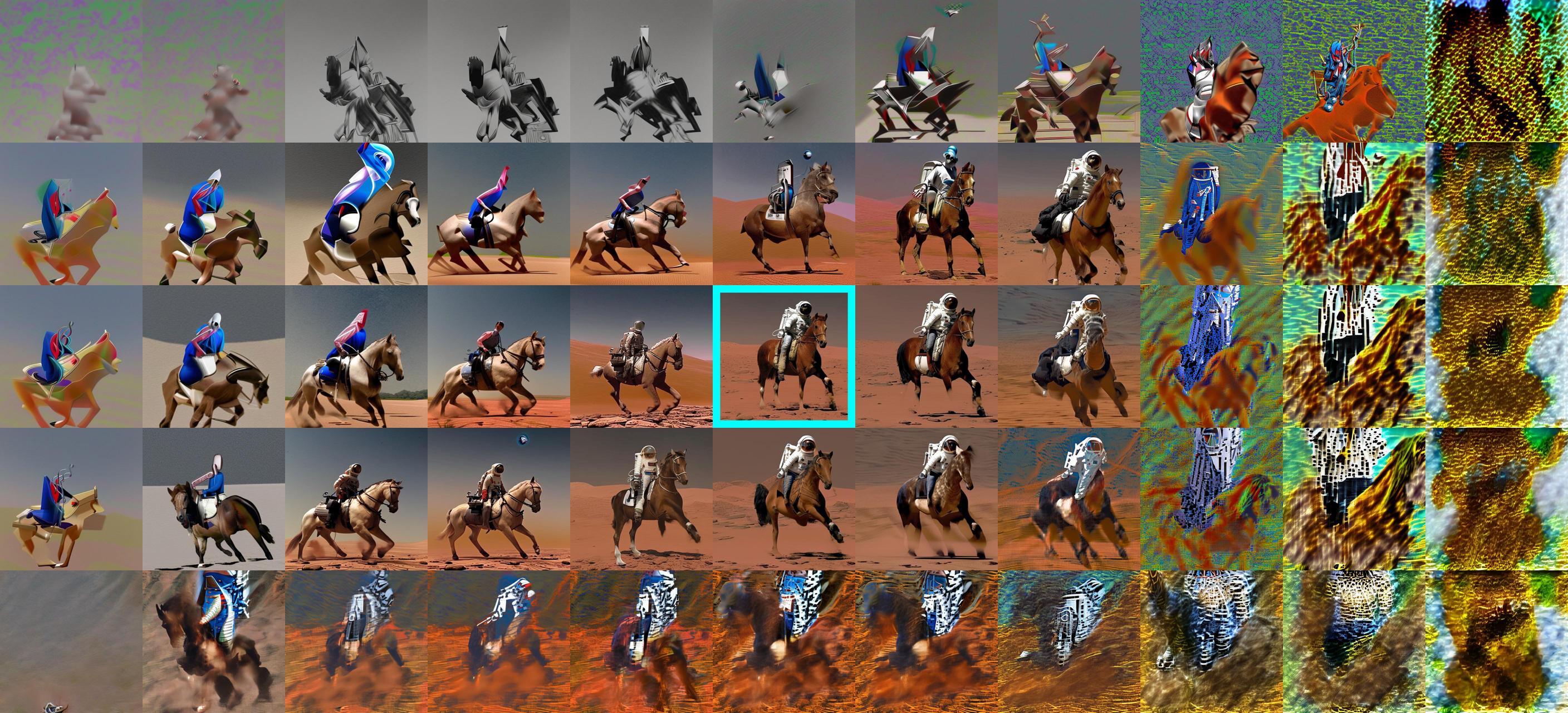}
        \includegraphics[width=1.0\textwidth]{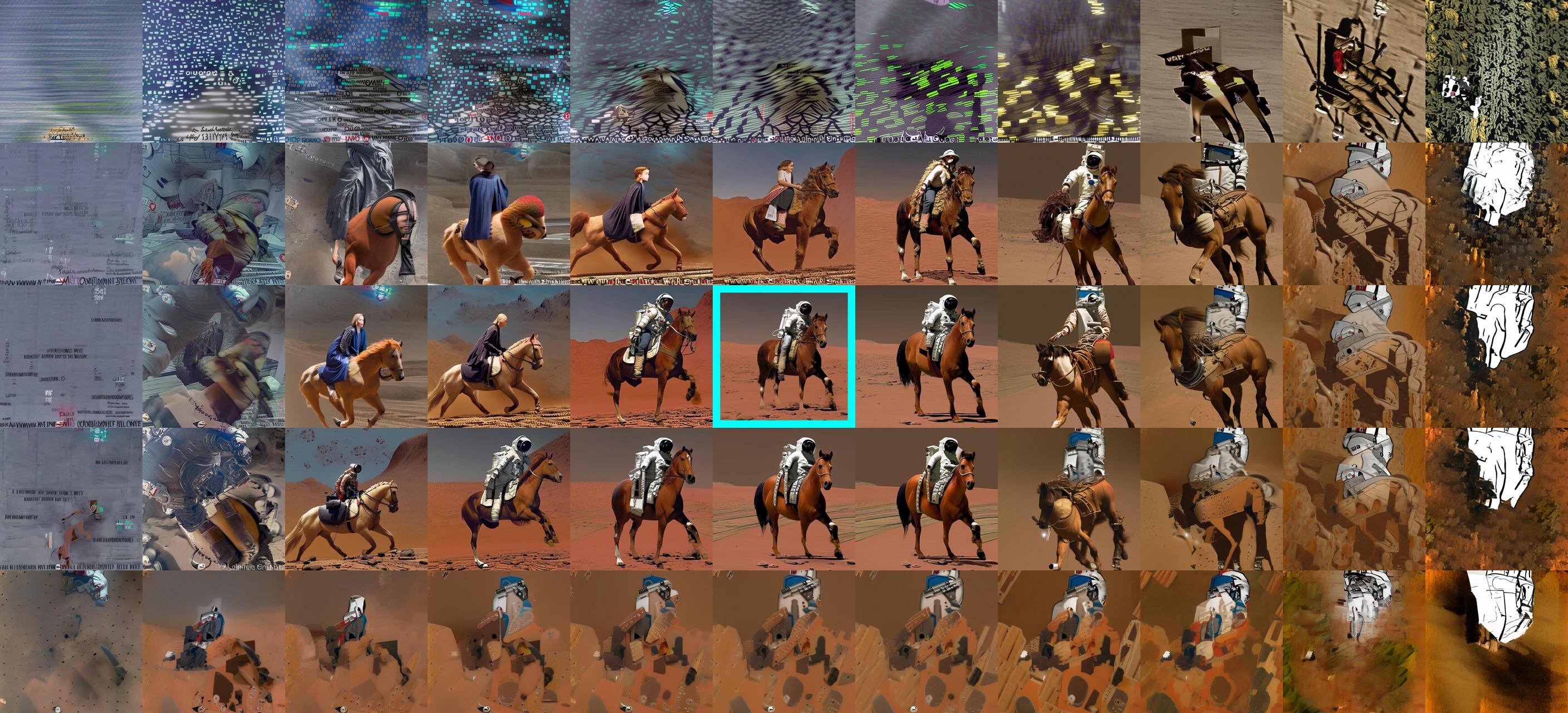}
        \includegraphics[width=1.0\textwidth]{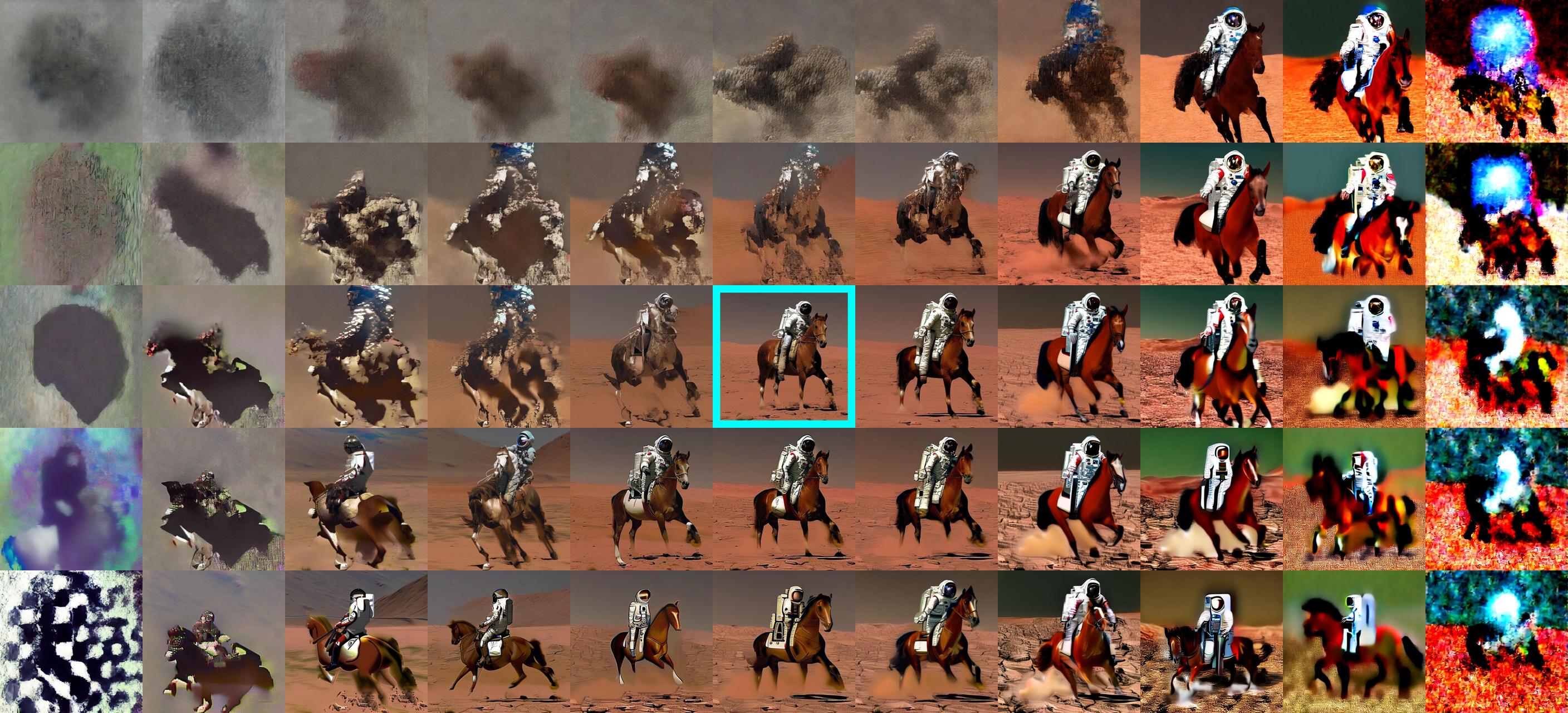}
    \end{subfigure}
    \caption{AttnMod scans applied to individual attention blocks. From top to bottom: U1A1A2, U1A0A2, and U2A0A1. Each grid is centered on the default output without AttnMod. Increasing the initial attention multiplier $\alpha$ moves rightward; increasing the in-loop change rate $\delta$ moves downward. All generations use the same prompt, seed, model, and scheduler settings as in Figures~\ref{fig:teaser} and~\ref{fig:picked_output}.}
    \label{fig:SD15_ARHoM_scans}
\end{figure}

Figure~\ref{fig:SD15_ARHoM_scans} shows example scans using Stable Diffusion 1.5 (SD1.5). Modulating the single attention block \texttt{"up\_blocks.1.attentions.1.transformer\_blocks.0.attn2.processor"}, abbreviated as \textbf{U1A1A2}, yields the top grid in Figure~\ref{fig:SD15_ARHoM_scans}. We vary $\alpha$ from $-10.0$ to $50.0$ and $\delta$ from $-1.0$ to $1.0$. The center tile represents the baseline image generated with $\alpha = 1.0$ and $\delta = 0.0$ (i.e., no modification). Moving horizontally increases $\alpha$, while moving vertically increases $\delta$, both in logarithmic scale. Additionally, Figure~\ref{fig:teaser} presents the scan for the \textbf{U1A2A2} block.

These scans reveal that manipulating attention produces diverse visual effects. Near the center of each grid, outputs remain faithful to the prompt, but farther from the center, the model begins to exhibit novel stylistic characteristics. Notably, even with negative attention scaling (e.g., $\alpha = -10.0$), the model still follows the prompt to a recognizable extent due to the remaining unmodified attention blocks.

The ability to independently modulate conditioning strength at different locations in the U-Net enables AttnMod to produce coherent, repeatable shifts in visual style. For instance, the far-left tile in the top row of the top grid in Figure~\ref{fig:SD15_ARHoM_scans} corresponds to the AttnMod setup \{\texttt{"U1A1A2"}: \{$\alpha$: $-10.0$, $\delta$: $-1.0$\}\}.

In the following sections, we explore how AttnMod can be applied across multiple blocks, evaluate the consistency of resulting styles, and analyze how styles persist across prompts and seeds.

\section{Qualitative Analysis}

\subsection{Seed Sensitivity}

\begin{figure}[ht]
    \centering
    \begin{subfigure}[b]{\textwidth}
        \includegraphics[width=1.0\textwidth]{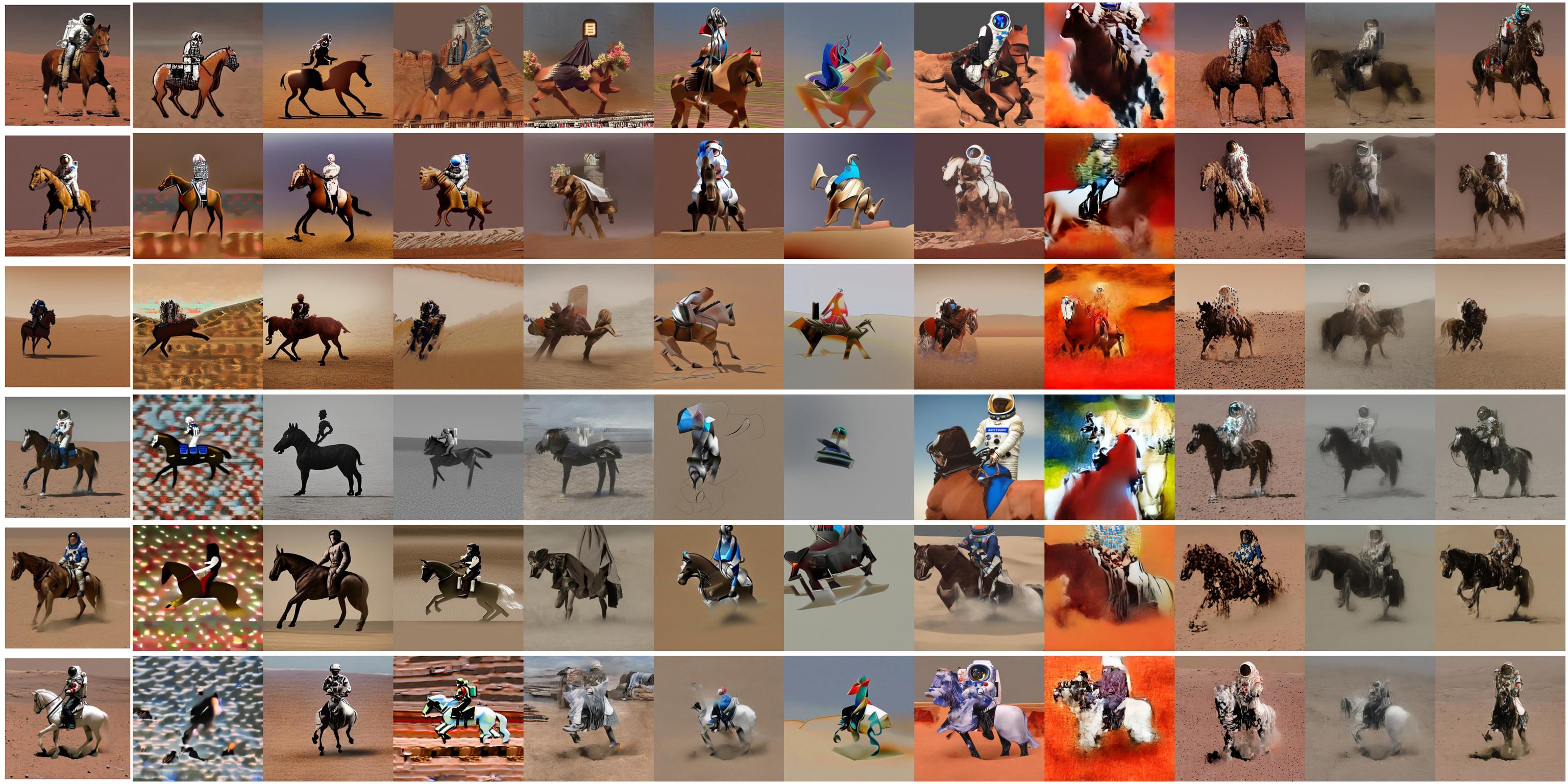}
    \end{subfigure}
    \caption{Effect of random seed on AttnMod outputs with all other inputs held constant. Each row corresponds to a different seed. Each column corresponds to a specific AttnMod setup. Images in the leftmost column are the default diffusion outputs without AttnMod.}
    \label{fig:SD15_ARHoM_seeds}
\end{figure}

To investigate how the random seed influences image generation under AttnMod, we conducted multiple single-block scans using the constant-rate setup described earlier. For each scan, we fixed all generation parameters except the seed, and applied the same AttnMod setup --- defined by a specific $\alpha$, $\delta$, and attention block. The resulting images are shown in Figure~\ref{fig:SD15_ARHoM_seeds}, with rows representing different seeds and columns representing the same AttnMod setup across seeds.

We observe that the generated outputs are influenced jointly by the AttnMod setup and the baseline generation associated with the given seed (shown in the leftmost column). This suggests that certain seeds may be more favorable for stylization, as they provide a better visual layout for AttnMod to build upon. A practical strategy, therefore, is to first select a seed that produces a visually strong composition using the base model, and then apply AttnMod to explore stylistic variations.

\subsection{Prompted Styles and Emergent Stylization}

\begin{figure}[ht]
    \centering
    \begin{subfigure}[b]{0.496\textwidth}
        \includegraphics[width=\textwidth]{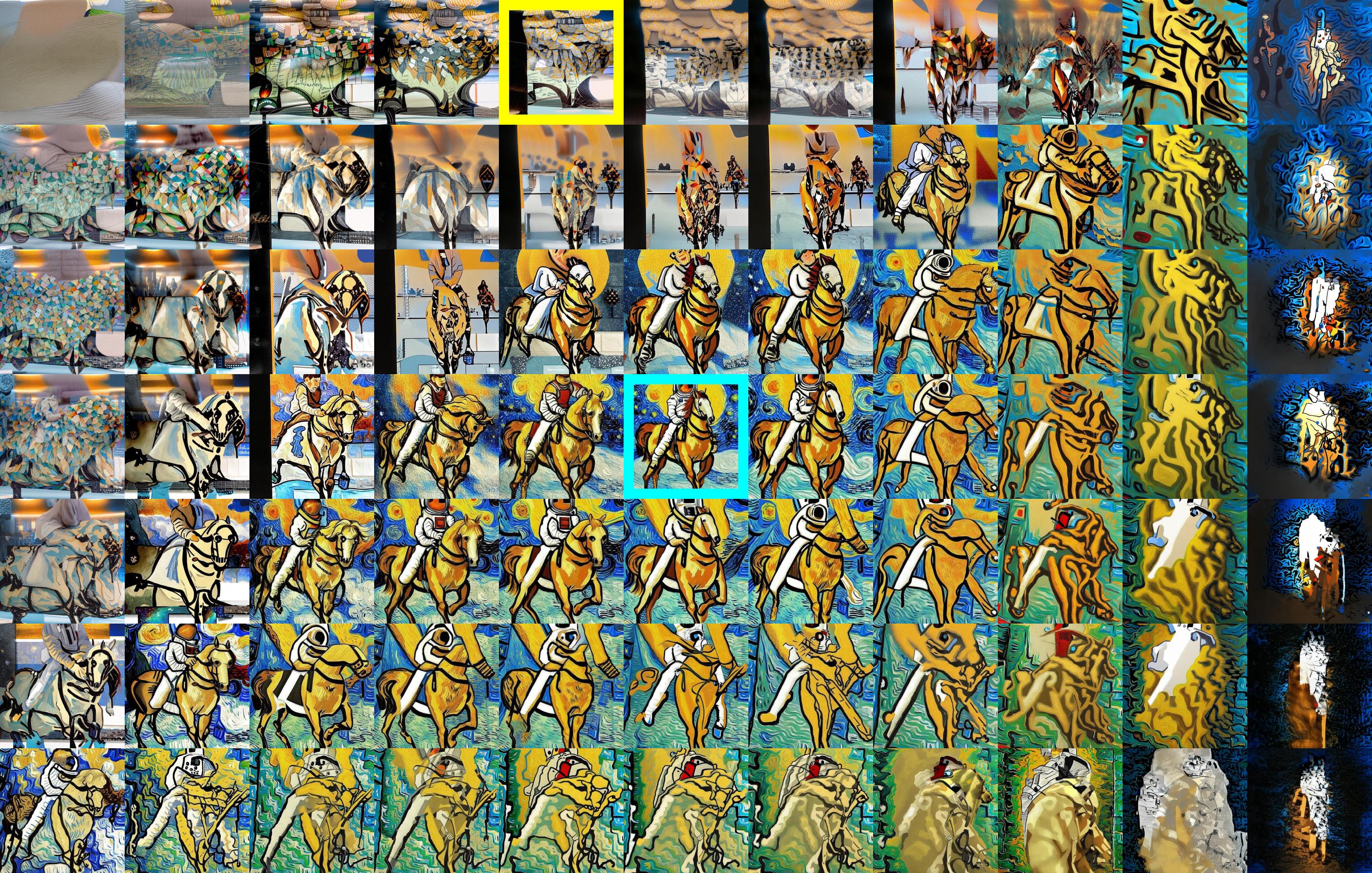}
        \caption{Prompted style: \textit{van Gogh}}
    \end{subfigure}
    \begin{subfigure}[b]{0.496\textwidth}
        \includegraphics[width=\textwidth]{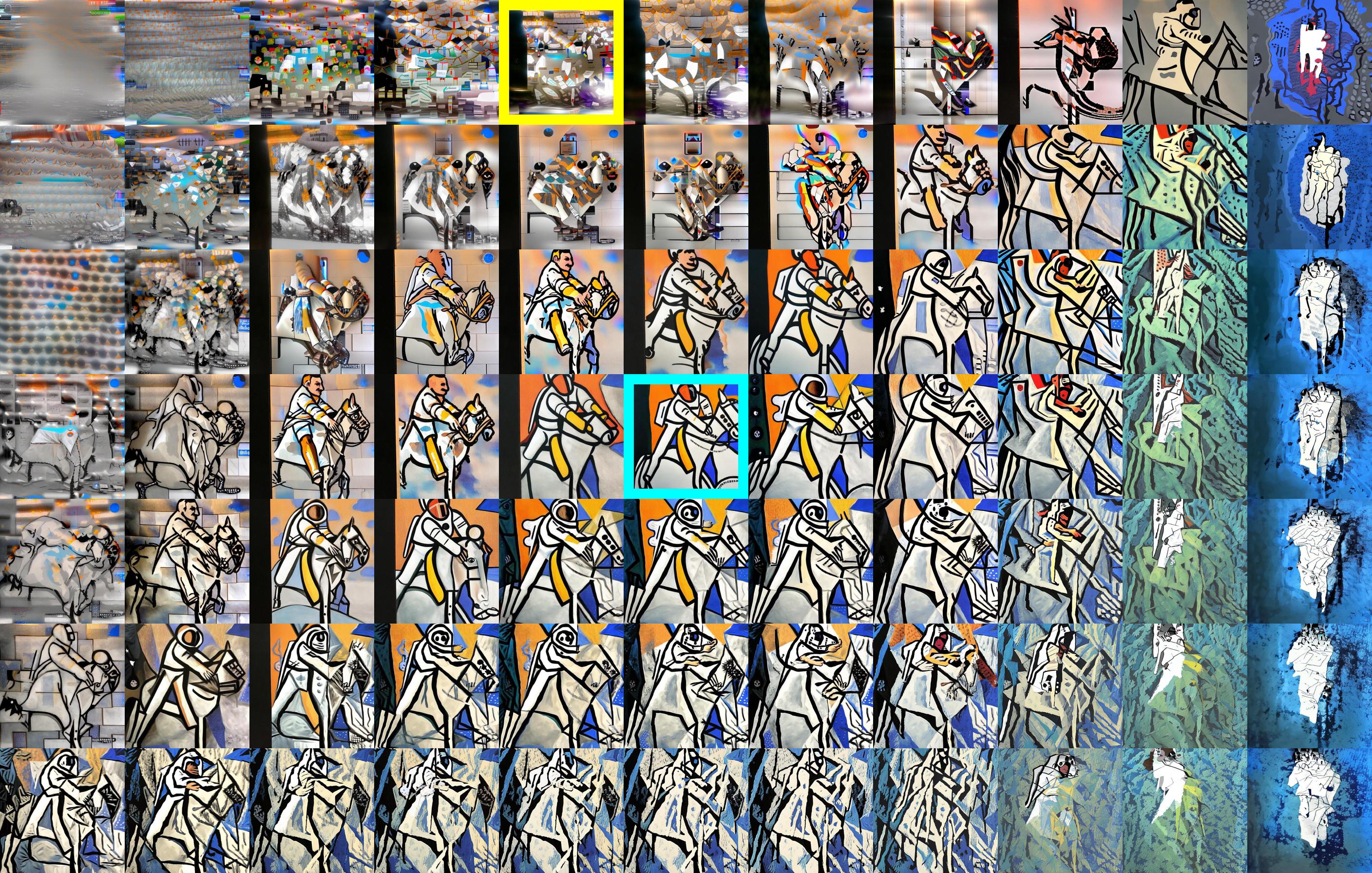}
        \caption{Prompted style: \textit{Picasso}}
    \end{subfigure}
    \begin{subfigure}[b]{\textwidth}
        \centering
        \includegraphics[width=\textwidth]{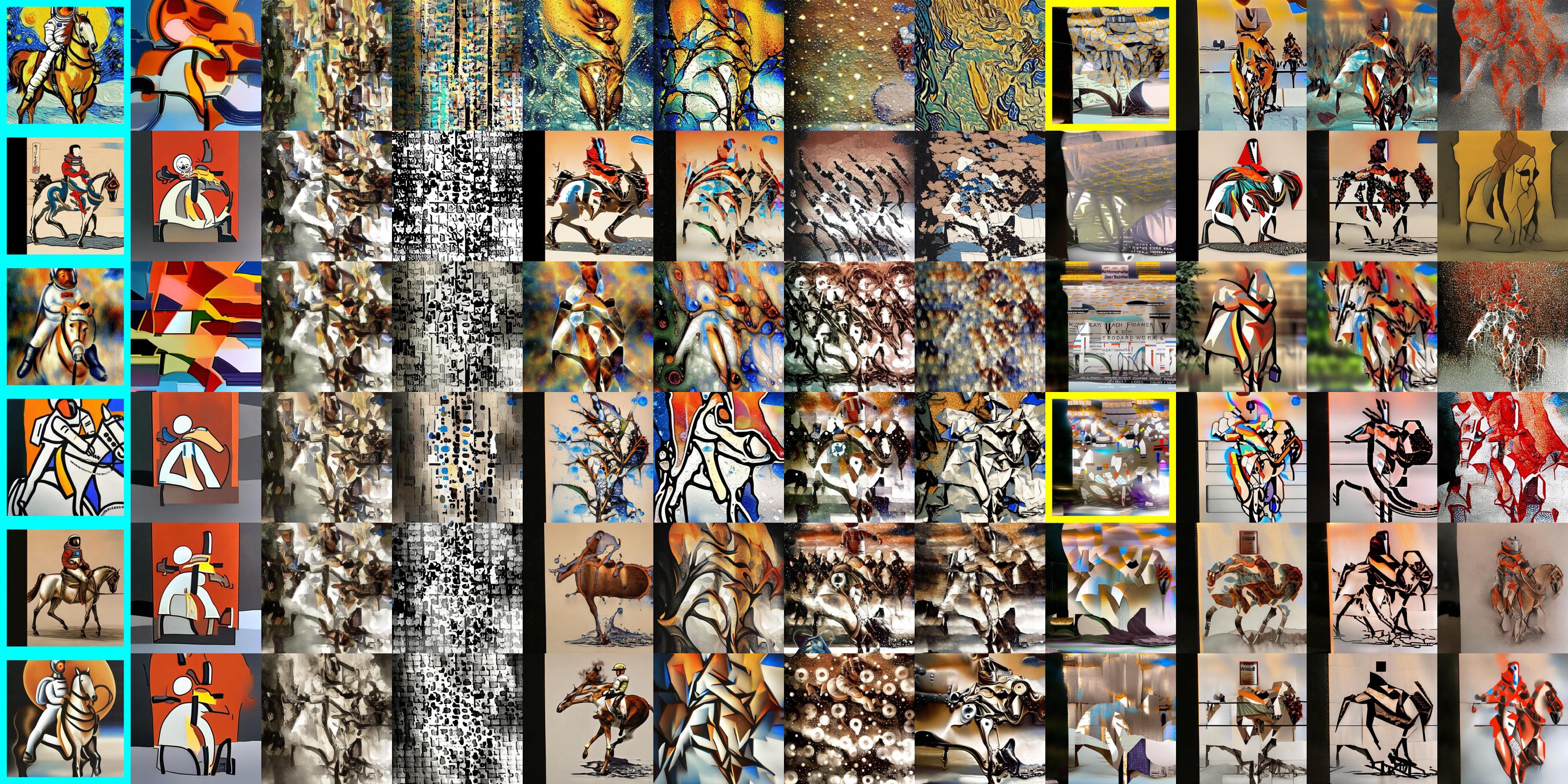}
        \caption{Emergent styles derived from AttnMod configurations, with columns corresponding to different setups. Yellow-highlighted tiles in (a) and (b) also appear here.}
    \end{subfigure}
    \caption{
    AttnMod applied to different prompted styles. (a) and (b) show how strong attention modification leads to new styles, distinct from the original prompts. (c) collects such emergent styles across prompts, with each column showing a different AttnMod configuration.
    }
    \label{fig:attnmod}
\end{figure}

To examine how AttnMod interacts with stylistic prompts, we apply single-block (\textbf{U1A0A2}) AttnMod to two example prompts: \textit{van Gogh} and \textit{Picasso}. Figures~\ref{fig:attnmod}(a) and~(b) show the resulting grids. The center tile in each grid represents the baseline output without attention modification. Tiles farther from the center reflect stronger AttnMod effects, which gradually reduce the influence of the original prompt.

At greater modification strengths, the generated images deviate significantly from the prompted style and begin to exhibit consistent, novel patterns. For instance, the yellow-highlighted tiles in Figures~\ref{fig:attnmod}(a) and~(b) share similar visual traits despite originating from different prompts, indicating the emergence of a new, AttnMod-driven style.

Figure~\ref{fig:attnmod}(c) collects such emergent styles across various prompts. The leftmost column shows unmodified outputs for style prompts: \textit{van Gogh}, \textit{Ukiyo-e}, \textit{Renoir}, \textit{Picasso}, \textit{da Vinci}, and \textit{Dali}. Each subsequent column applies a distinct AttnMod configuration while keeping the seed and diffusion parameters fixed. These results highlight how AttnMod can steer the generation toward new styles that are only partially influenced by the text prompt.

\section{Conclusion}

We introduce AttnMod, a simple yet powerful method for steering the visual style of diffusion-generated images by modulating cross-attention strength during the denoising process. AttnMod operates without modifying model weights, and works seamlessly across pretrained models such as SD1.5, SDXL, and their derivatives.

Our experiments demonstrate that AttnMod can produce coherent, prompt-aligned images with novel stylistic characteristics—even in cases where the desired style is difficult to express through text alone. By treating attention as a tunable stylistic channel, AttnMod enables creators to explore a broader visual space with minimal intervention.

Looking ahead, future work may explore richer strategies for multi-block coordination, adaptive in-loop modulation, and integration with existing fine-tuning or personalization techniques.

\bibliographystyle{unsrt}
\bibliography{ref}

\appendix

\section*{Appendix A: Ablation Study}

We observe that disabling cross-attention at a single block (i.e., $\alpha = 0.0$, $\delta = 0.0$) results in minimal change to the generated image compared to the default diffusion output. This suggests that removing prompt conditioning alone is insufficient to produce stylistic shifts.

To induce a noticeable change in style, either a stronger initial attention multiplier ($|\alpha| > 2.0$) or a nontrivial in-loop drift ($|\delta| > 0.2$) is typically required, assuming a 30-step denoising schedule. These thresholds serve as practical baselines when selecting AttnMod setups for stylization.

\section*{Appendix B: Multi-Block Attention Strategies}

\begin{figure}[ht]
    \centering
    \begin{subfigure}[b]{0.39\textwidth}
        \includegraphics[width=\textwidth]{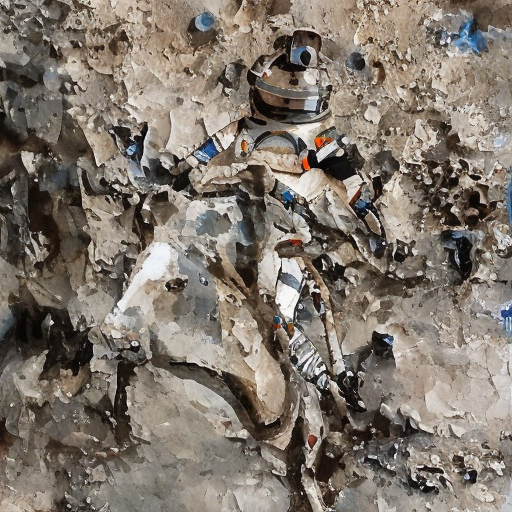}
        \caption{Most influential block first}
    \end{subfigure}
    \hspace{0.1\textwidth}
    \begin{subfigure}[b]{0.39\textwidth}
        \includegraphics[width=\textwidth]{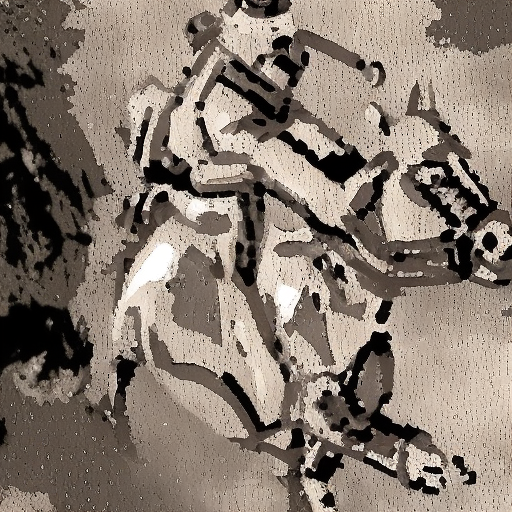}
        \caption{Least influential block first}
    \end{subfigure}
    \caption{Two multi-block AttnMod strategies using all 32 attention blocks in SD1.5, each begins with zero attention across all blocks and activates one per denoising step. The only difference is the block selection criterion: (a) maximizes per-step visual change, while (b) minimizes it.}
    \label{fig:MultiBlocks}
\end{figure}

Stable Diffusion 1.5 contains 32 modifiable cross-attention blocks within the U-Net. Exploring combinations of these blocks for attention modulation yields a vast design space. Figure~\ref{fig:MultiBlocks} presents a controlled experiment over this space, where all 32 blocks are initialized with zero attention. During each of the 40 denoising steps, one block is selected to receive an increment of attention up to a maximum of 1.0. This defines a 40-step modulated denoising strategy with a sparse attention budget.

Two block selection heuristics are compared. The first selects, at each step, the available block whose activation produces the greatest perceptual difference relative to the previous image. This favors aggressive conditioning and aims to amplify stylistic transformation. The second strategy does the opposite, selecting the block that induces the least change, thus attempting to preserve ambiguity or allow style to emerge more gradually. Both approaches result in visually distinctive outcomes and demonstrate that attention scheduling—across both space and time—offers a rich axis for generating new, unpromptable styles. Further refinements or hybrid scheduling strategies remain promising directions for future exploration.

\section*{Appendix C: Objects and Checkpoints}

\begin{figure}[ht]
    \centering
    \begin{subfigure}[b]{\textwidth}
        \includegraphics[width=1.0\textwidth]{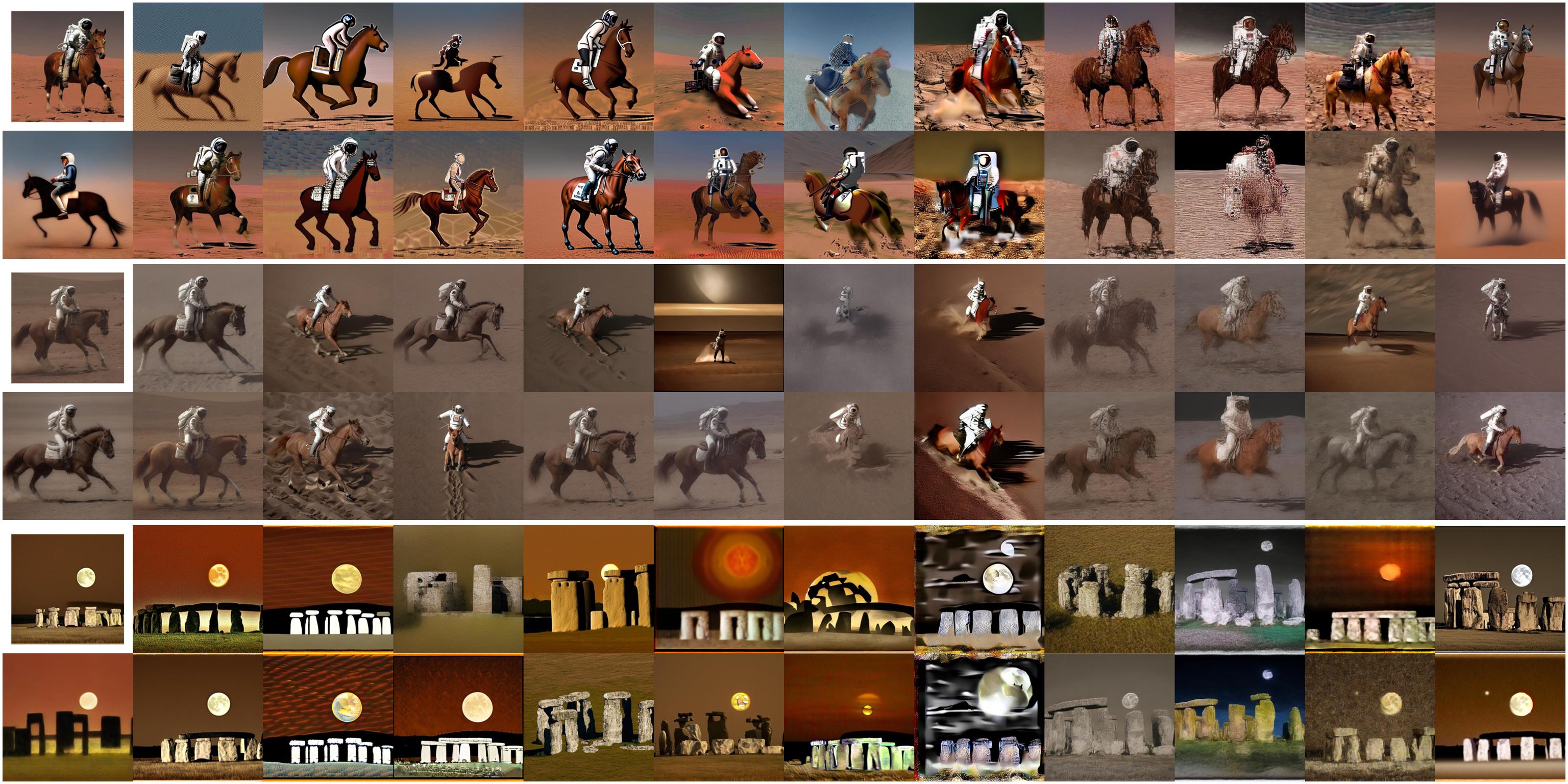}
    \end{subfigure}
    \caption{AttnMod applied across different checkpoints and prompts. The top and middle sections use the same prompt but different diffusion checkpoints (SD1.5 and RealisticVisionV5.1~\cite{SG1612222023}). The top and bottom sections use the same SD1.5 checkpoint but different prompts.}
    \label{fig:PCP}
\end{figure}

To evaluate AttnMod's behavior across different model checkpoints and scene prompts, Figure~\ref{fig:PCP} presents three parallel generations. The top and middle sections use the same text prompt but different checkpoints: the top uses the official SD1.5 model, while the middle uses RealisticVisionV5.1~\cite{SG1612222023}. Despite using the same AttnMod parameters, the resulting styles differ, highlighting how checkpoint architecture and training biases influence the stylization effect.

The bottom section reuses the SD1.5 checkpoint from the top but applies a different prompt describing a distinct scene. Since AttnMod modifies cross-attention based on prompt embeddings, the resulting styles can shift accordingly. Some setups remain visually consistent across scenes, while others adapt to scene content, producing new yet coherent styles.

These results suggest that both the base model and the semantic content of the prompt modulate how AttnMod expresses stylization, reinforcing its flexibility across content and architecture domains.

\end{document}